\documentclass{article}
\usepackage{spconf,amsmath,epsfig}
\usepackage{amssymb,amsfonts,url}
\usepackage{subfigure, multirow}
\renewcommand{\vec}[1]{\mathbf{#1}}

\begin{document}\sloppy

\def\x{{\mathbf x}}
\def\L{{\cal L}}

\title{Feature Preserving and Uniformity-Controllable Point Cloud Simplification on Graph}
%
\name{Junkun Qi, Wei Hu, Zongming Guo}
\address{Institute of Computer Science \& Technology, Peking University}

\maketitle

\begin{abstract}
With the development of 3D sensing technologies, point clouds have attracted increasing attention in a variety of applications for 3D object representation, such as autonomous driving, 3D immersive tele-presence and heritage reconstruction.
However, it is challenging to process large-scale point clouds in terms of both computation time and storage due to the tremendous amounts of data.
Hence, we propose a point cloud simplification algorithm, aiming to strike a balance between preserving sharp features and keeping uniform density during resampling. In particular, leveraging on graph spectral processing, we represent irregular point clouds naturally on graphs, and propose concise formulations of feature preservation and density uniformity based on graph filters. The problem of point cloud simplification is finally formulated as a trade-off between the two factors and efficiently solved by our proposed algorithm. Experimental results demonstrate the superiority of our method, as well as its efficient application in point cloud registration.
\end{abstract}

\section{Introduction}
\label{sec:intro}
\begin{figure}[t]
\begin{minipage}[b]{0.45\linewidth}
  \centering
  \centerline{\epsfig{figure=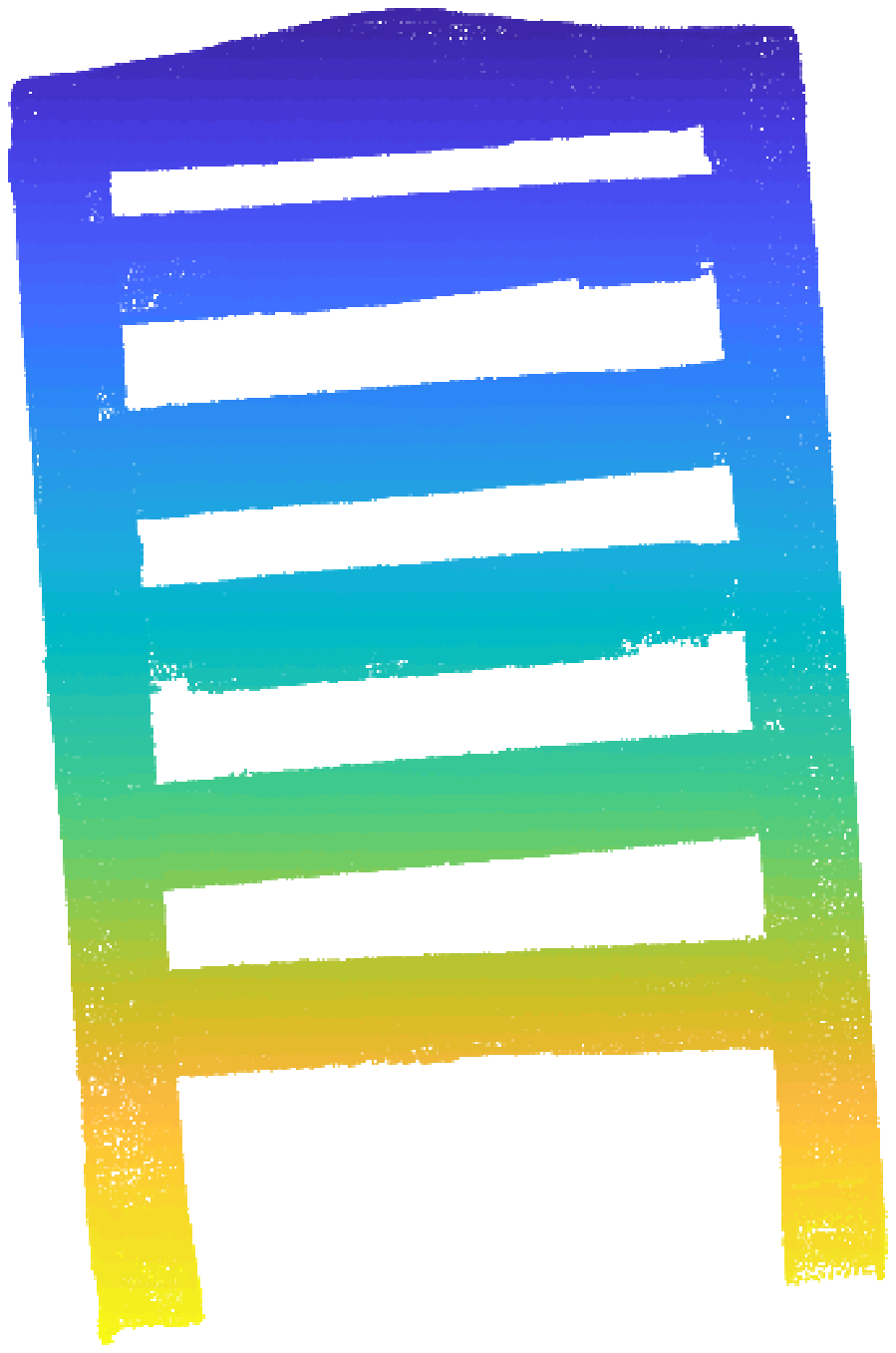,width=4.0cm}}
  \vspace{0.1cm}
  \centerline{(a) Original point cloud}\medskip
\end{minipage}
\hfill
\begin{minipage}[b]{0.45\linewidth}
  \centering
  \centerline{\epsfig{figure=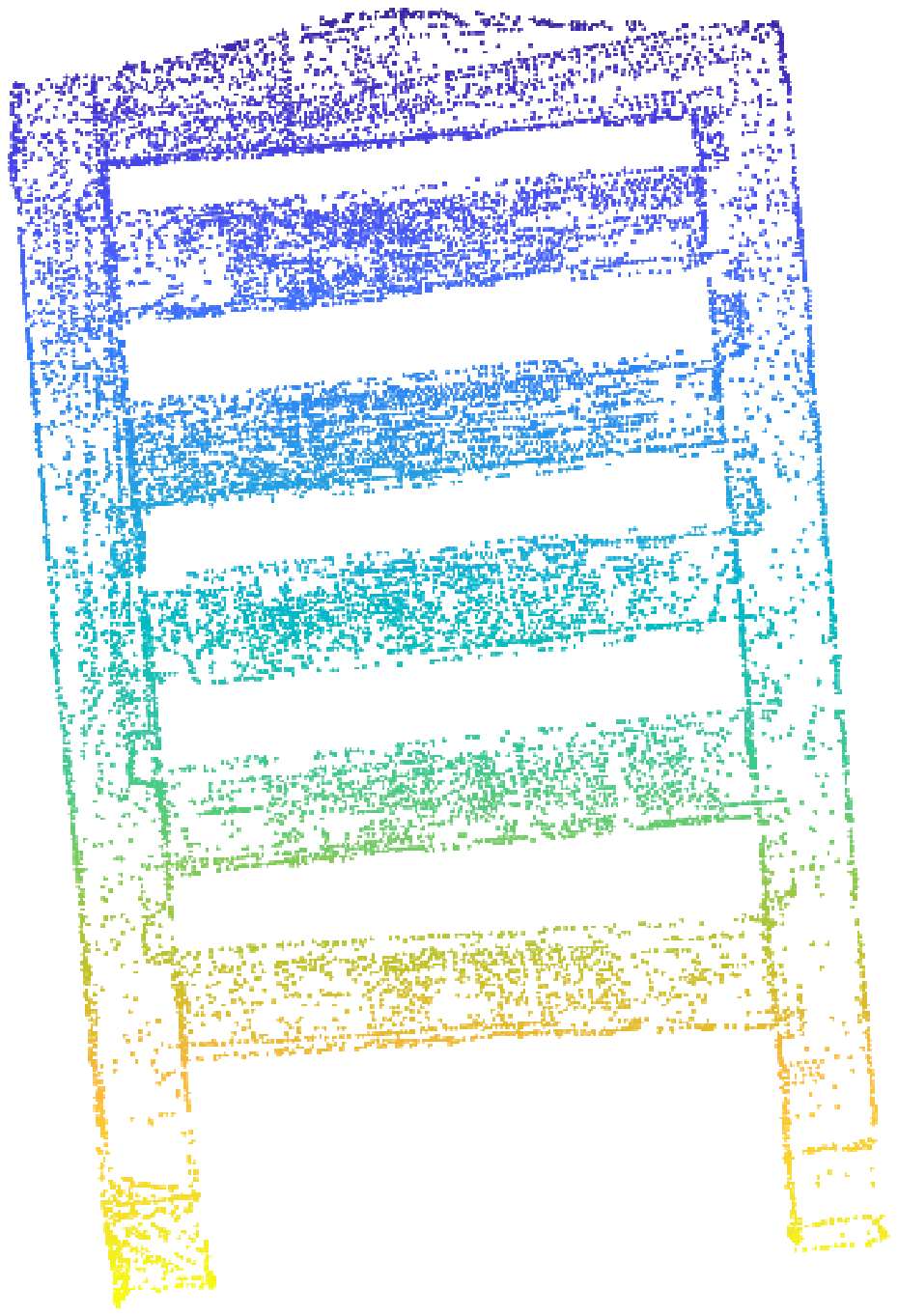,width=4.0cm}}
  \vspace{0.1cm}
  \centerline{(b) Simplified point cloud}\medskip
\end{minipage}
\vspace{-0.1in}
\caption{The proposed point cloud simplification method enhances contours of the point cloud while retaining the density uniformity. (a) shows the original point cloud of \textit{Shutter}, which consists of 291,220 points. (b) shows the simplified point cloud with 5\% of points reserved. }
\label{fig:shutter_example}
\end{figure}

The development of 3D sensing technologies enables the convenient acquisition of large-scale point clouds.
A point cloud is a natural representation of arbitrarily-shaped objects, which consists of a set of points on irregular domain.
Each point has 3D coordinates and possibly other attributes such as color and normal.
Point clouds have been widely applied in various fields, such as 3D immersive tele-presence, navigation for autonomous driving, and heritage reconstruction \cite{tulvan16use}.
Nevertheless, it is challenging to process \textit{large-scale} point clouds in terms of both computation time and storage due to the tremendous amounts of data. Hence, point cloud simplification (or resampling, downsampling) is required.

Existing point cloud simplification algorithms can be mainly classified into two types: mesh-based simplification and point-based simplification.
Earlier algorithms are based on mesh reconstructed from the point cloud, which contains not only points but also surfaces \cite{cignoni1998comparison}.
However, the process of mesh reconstruction is quite time-consuming for large-scale point clouds, and thus point-based simplification is proposed.
Point-based simplification makes use of the information of the raw points to determine whether a point is to be preserved or abandoned, and sometimes new points are generated.
Note that, both types of algorithms are heuristic without optimization, and often lead to artifacts such as edge deficiencies.

Chen et al. \cite{chen2018fast} optimize the resampling distribution by minimizing the proposed reconstruction error based on a feature-extraction operator. The contours in the point clouds are well preserved after resampling, but the points are extremely nonuniform. A point cloud is regarded uniform if the local density of points is similar in different regions. In practice, the uniformity property of point clouds is often desired to facilitate applications such as rendering, denoising, inpainting, etc. Therefore, \textit{a good balance between feature preserving and uniformity is in need}.  

Hence, we propose an \textit{optimized} point cloud simplification approach, which is optimal in terms of striking a balance between the preservation of \textit{sharp features} and \textit{the uniformity} of the point cloud, as demonstrated in Fig.~\ref{fig:shutter_example}. We cast this goal as an optimization problem, with a user-adjustable parameter to control the degree of uniformity for various applications. In particular, we represent point clouds naturally on graphs, with each point as a vertex in the graph and the relationship among points described by edges. Based on the representation, we propose concise formulation of the loss in feature via high-pass graph filters and the loss in density uniformity via the number of graph connectivities of each vertex, leveraging on the field of graph signal processing \cite{shuman2013emerging}. Further, we propose constraint relaxation and an efficient algorithm to solve the formulated optimization problem.

In summary, our contributions include:
\begin{itemize}
    \item We propose optimized point cloud simplification, aiming to strike a balance between feature preservation and density uniformity. 
    \item We formulate the loss in feature preservation and density uniformity concisely, leveraging on graph signal processing.
    \item Experimental results validate the superiority of our method, as well as the effectiveness as a preprocessing step for the application of point cloud registration.  
\end{itemize}

The paper is organized as follows.
We first review previous point-based simplification algorithms in Sec.~\ref{sec:related} and basic concepts of graph signal processing in Sec.~\ref{sec:background}.
Then we state the problem and formulate it as an optimization problem in Sec.~\ref{sec:formulation}.
An efficient algorithm to solve the optimization problem is proposed in Sec.~\ref{sec:algorithm}.
Finally, experimental results and conclusions are discussed in Sec.~\ref{sec:results} and Sec.~\ref{sec:conclude}, respectively.

\section{Related Work}
\label{sec:related}
We briefly review existing point-based simplification algorithms, including clustering-based, iteration-based and formulation-based simplification.

\textbf{Clustering-based simplification}~~
The idea is to divide the point cloud into clusters and then replace the points in each cluster by one or several points.
Yu et al. \cite{yu2010asm} cluster the points by hierarchical clustering followed by a local clustering to minimize the sample error.
Shi et al. \cite{shi2011adaptive} use the $K$-means clustering method, detect the boundary clusters and subdivide the cluster with high curvature.
As clustering large amounts of points is time-consuming, Benhabiles et al. \cite{benhabiles2013fast} propose coarse-to-fine approach and create a coarse cloud using volumetric clustering approach to speed up the algorithm. In general, clustering-based simplification is amenable to implementation, but usually causes artifacts such as edge deficiencies.

\textbf{Iteration-based simplification}~~
Moenning et al. \cite{moenning2004intrinsic} introduce the farthest point resampling method, selecting the points iteratively according to the Voronoi diagrams.
Song et al. \cite{song2009progressive} take advantage of the normals and distances of the neighbors of one point to define the importance of the point.
Then they remove the point with the least significance and update the significance of the remaining points iteratively.
Lee et al. \cite{lee2011dso} also use normals to define the importance of each point, but they merge two points with the least significance instead in each step.
Yang et al. \cite{yang2015feature} define the mean curvature of points by Principal Component Analysis and Fourier Transform, and iteratively remove the points around the point with the largest curvature in the remaining ones.

Iterative-based simplification often obtains better performance than clustering-based methods, but it is less efficient in the process of updating and finding points with the largest (or least) significance after each step.

\textbf{Formulation-based simplification}~~
Both clustering-based and iterative-based simplification have no proof of optimality.
In order to be more mathematically rigorous, Leal et al. \cite{leal2017linear} cluster the points and then identify points with high curvature which will be preserved.
For the remaining points, they use the linear programming model to select a reduced set with density equivalent to the original data set.
Chen et al. \cite{chen2018fast} define a resampling distribution and simplify the point cloud according to the distribution.
The optimized distribution is acquired by minimizing the proposed reconstruction error based on the proposed feature-extraction operator. While the contours are thus well preserved, the resulting point cloud is extremely nonuniform, which might be a hurdle to some applications such as rendering, denoising, etc. This motivates us to propose a simplification method that minimizes the loss in both feature preservation and density uniformity. 

\section{Background}
\label{sec:background}
We address point cloud simplification leveraging on graph signal processing. The basic concepts of spectral graph theory~\cite{chung1997spectral} are reviewed here, including graph, graph Laplacian and graph signal.

We consider an undirected graph $ \mathcal{G}=\{\mathcal{V},\mathcal{E},\mathbf{W}\} $ composed of a vertex set $ \mathcal{V} $ of cardinality $|\mathcal{V}|=N$, an edge set $ \mathcal{E} $ connecting vertices, and a weighted adjacency matrix $ \mathbf{W} $. $ \mathbf{W} $ is a real symmetric $ N \times N $ matrix, where $ W_{i,j} $ is the weight assigned to the edge $ (i,j) $ connecting vertices $ i $ and $ j $. We assume non-negative weights, i.e. $W_{i,j} \geq 0$. For instance, the graph adopted in our work is a $k$-nearest-neighbor ($k$-NN) graph, where each vertex is connected to its $k$ nearest neighbors.  

The graph Laplacian matrix, defined from the adjacency matrix, can be used to uncover many useful properties of a graph. Among different variants of Laplacian matrices, the combinatorial graph Laplacian used in \cite{shen2010edge,hu2012depth,hu2015multiresolution} is defined as $ \mathcal{L}=\mathbf{D}-\mathbf{W} $, where $ \mathbf{D} $ is the degree matrix--a diagonal matrix where $ D_{i,i} = \sum_{j=1}^N W_{i,j} $. Further, the  graph Laplacian can be normalized as $ \mathbf{L} = \mathbf{D}^{-1} \mathcal{L} = \mathbf{I} - \mathbf{D}^{-1} \mathbf{W} $.

Graph signal refers to data residing on the vertices of a graph, such as social, transportation, sensor, and neuronal networks. In our context, we construct a $k$-NN graph on the point cloud, where the coordinate of each point can be treated as the graph signal defined on the $k$-NN graph. This will be discussed further in Sec.~\ref{sec:formulation}.

\section{Problem Formulation}
\label{sec:formulation}
We describe point cloud simplification as a process of resampling the point cloud:
given a point cloud $\mathbf{X}$ with $|\mathbf{X}| = N$, find a point cloud $\mathbf{X}' \subset \mathbf{X}$
with $|\mathbf{X}'| = M < N$. The simplification rate is defined as $\alpha = \frac M N$.

A point cloud $\mathbf{X}$ with $N$ points, each of which is composed of $K$ attributes, is represented as
$\mathbf{X} \in \mathbb R^{N \times K}$, where the $i$-th row denoted as $\vec{x}_i ^ {\mathrm{T}}$
represents the $i$-th point.
Attributes can be coordinates, normals, colors, etc., where coordinates are compulsory, i.e., $K \ge 3$.
In order to represent the simplified point cloud, we introduce a resampling diagonal matrix
denoted as $\mathbf{\Psi} \in \mathbb R^{N \times N}$ with $\Psi_{i,i} = 1$
if $\vec{x}_i$ is kept in the simplified point cloud and $\Psi_{i,i} = 0$ otherwise. 
Thus, the simplified point cloud is represented as $\mathbf{\Psi X}$.

Our goal is to find the optimal resampling matrix $\mathbf{\Psi}$ so as to keep most geometry
features (e.g., contours of the point cloud) while controlling its uniformity. This is essentially \textit{a trade-off between the uniformity of points in the simplified point cloud and the preservation of sharp features}. Hence, we first formulate the loss in feature and the loss in uniformity due to simplification respectively. 
Then we cast the problem of finding the optimal resampling matrix as an optimization problem, which minimizes the total loss.

\subsection{Loss in Feature}
\label{subsec:loss_feature}
Inspired by \cite{chen2018fast}, we leverage the normalized graph Laplacian $ \mathbf{L} $---a high-pass filter---to extract sharp features of the graph signal. We firstly construct a $k$-NN graph on the point cloud. The attribute of the point cloud is then regarded as the graph signal.
For simplicity, we assume the attribute consists of merely coordinates. We then define the edge weight $W_{i,j}$ between vertices $i$ and $j$ as an exponential function of the Euclidean distance between $i$ and $j$:
\begin{equation}
W_{i,j} = 
\begin{cases}
\exp\left({- \dfrac {\|\vec{x}_i-\vec{x}_j\|^2_2} {\sigma^2} }\right), & j \in \mathcal{N}_i\\
0, & \text{otherwise}
\end{cases}
\label{eq:weight}
\end{equation}
where $\sigma$ is a parameter, and $\mathcal{N}_i$ denotes the set of neighbors of vertex $i$.

We denote the normalized edge weight between $\vec{x}_i$ and $\vec{x}_j$ as $\widetilde{W}_{i,j}=\frac {W_{i,j}} {\sum\limits_{j} W_{i,j}}$,
where $\widetilde{\mathbf{W}} = \mathbf{D}^{-1} \mathbf{W}$ is the normalized weight matrix.
Then the $i$-th row of the matrix $\mathbf{L X}$, $\widetilde{\mathbf{X}}_i$, follows as  
\begin{equation}
(\mathbf{L X})(i)=\widetilde{\mathbf{X}}_i = \vec{x}_i - \sum\limits_{j} \widetilde{W}_{i,j} \vec{x}_j.
\end{equation}
As defined in Eq.~\ref{eq:weight}, the edge weight encodes the similarity between two points. Hence, $\widetilde{\mathbf{X}}_i$ encodes the variation of one point from its neighbors. This is because $\sum\limits_{j} \widetilde{W}_{i,j} \vec{x}_j$ is a weighted representation of $\vec{x}_i$'s neighbors, which would differ from $\vec{x}_i$ greatly if $\vec{x}_i$ is distinct from its neighbors, resulting in large $\|\widetilde{\mathbf{X}}_i\|_2$.
As we know, points that are distinct from its neighbors tend to exhibit sharp features, such as points on a contour.
Accordingly, a large $\|\widetilde{\mathbf{X}}_i\|_2$ is likely to correspond to sharp features.

Hence, we represent sharp features of the point cloud $\mathbf{X}$ as $\mathbf{L X}$, and the remaining features after resampling as $\mathbf{\Psi L X}$ for simplicity.
The loss in feature due to simplification is thus defined as
\begin{equation}
l_f(\mathbf{\Psi}) = \|\mathbf{\Psi L X} - \mathbf{L X}\|^2_2.
\end{equation}

\subsection{Loss in Density Uniformity}
While deploying the normalized Laplacian is able to preserve sharp features well, as in \cite{chen2018fast}, points on surfaces with indistinct features will almost be all neglected, leading to extreme density
non-uniformity of the point cloud, i.e., with almost only contours remaining after simplification.
In order to avoid this extreme non-uniformity, we further define loss in the density uniformity for regularization, leveraging the degree of each vertex.  

As in Sec.~\ref{subsec:loss_feature}, we construct a $k$-NN graph. If the density of the original point cloud is uniform, the $k$ nearest neighbors lie in a ball centering at each point with the same radius. We use a binary matrix $\mathbf{A}$ to represent the adjacency of the graph, i.e. $A_{i,j} = 1$ if and
only if $\vec{x}_j$ is one neighbor of $\vec{x}_i$.
Each row of $\mathbf{A}$ indicates the neighbors of a point, and sums up to $k$.
By means of the definition of $\mathbf{\Psi}$, we represent the adjacency matrix of the simplified point cloud graph as $\mathbf{A \Psi}$.
Given the simplification rate $\alpha$, the number of each point's neighbors in the graph constructed over the simplified point cloud is approximately equal to $\alpha k$ if simplified uniformly.
Hence, we define the uniformity loss as
\begin{equation}
l_e(\mathbf{\Psi}) = \| \mathbf{A \Psi} \vec{1} - \alpha k \vec{1} \|^2_2,
\end{equation}
where $\vec{1} \in \mathbb{R}^N$ represents a column vector with every element equal to $ 1 $. Thus $\mathbf{A \Psi} \vec{1}$ computes the number of neighbors of each point in the simplified point cloud.  

\subsection{Final Objective}

Having defined the loss in feature and uniformity, we formulate point cloud simplification as an optimization problem, in which the objective function aims to strike a balance between the feature loss and the uniformity loss:
\begin{equation}
l(\mathbf{\Psi}) = l_f(\mathbf{\Psi}) + \lambda l_e(\mathbf{\Psi}),
\end{equation}
where $\lambda$ is a hyper-parameter to keep a balance of the feature and uniformity.
Further, we add some constraints for the optimization variable $ \mathbf{\Psi} $ to make it valid. 
The problem formulation is
\begin{equation}
\begin{split}
\min\limits_{\mathbf{\Psi}} \ & \| \mathbf{\Psi L X} - \mathbf{L X} \|^2_2 
+ \lambda \| \mathbf{A \Psi} \vec{1} - \alpha k \vec{1}\|^2_2,\\
\text{s.t.} \ & \Psi_{i,i} \in \{0,1\}, \ i = 1,2,...,N;\\
& \Psi_{i,j} = 0, \ i \neq j;\\
& tr(\mathbf{\Psi}) = \alpha N.
\end{split}
\label{eq:original}
\end{equation}

In order to facilitate solving the optimization problem, we define a resampling vector
$\vec{\psi} \in \mathbb R^N$.
$\vec{\psi}$ is actually the diagonal elements of the resampling matrix $\mathbf{\Psi}$, i.e. $\psi_i = \Psi_{i,i}, \ i = 1,2,...,N$.
We then replace $\mathbf{\Psi}$ with $\vec{\psi}$ in the optimization objective of (\ref{eq:original}).

Further, we define the feature matrix $(\mathbf{L X})(\mathbf{L X})^\mathrm{T}$ as $\mathcal{F}$ for simplicity, and define a diagonal matrix $\mathbf{F} \in \mathbb{R}^{N \times N}$ and a vector $\vec{f} \in \mathbb{R}^N$ with $f_i = F_{i,i} = \mathcal{F}_{i,i}$, $i = 1,2,...,N$.
Then the aforementioned problem formulation is equivalent to
\begin{equation}
\begin{split}
\min\limits_{\vec{\psi}} \ & \vec{\psi}^\mathrm{T} \mathbf{F} \vec{\psi} - 2 \vec{f}^\mathrm{T} \vec{\psi} +
\lambda \left[ \vec{\psi}^\mathrm{T} \mathbf{A}^\mathrm{T} \mathbf{A} \vec{\psi} - 2 \alpha k (\mathbf{A}
\vec{1})^\mathrm{T} \vec{\psi} \right]\\
\text{s.t.} \ & \psi_i \in \{0,1\}, \ i = 1,2,...,N;\\
& \vec{\psi}^\mathrm{T} \vec{1} = \alpha N.
\end{split}
\label{eq:final1}
\end{equation}

\section{The Proposed Algorithm}
\label{sec:algorithm}
The optimization problem in (\ref{eq:final1}) is a combinatorial optimization problem, which is NP-hard.
In order to solve the algorithm efficiently, we relax the first constraint to $0 \le \psi_{i} \le 1$.
Then the optimization problem is simplified to
\begin{equation}
\begin{split}
\min\limits_{\vec{\psi}}~ \ & \vec{\psi}^\mathrm{T} (\mathbf{F} + \lambda \mathbf{A}^\mathrm{T} \mathbf{A}) \vec{\psi} - 2(\vec{f} + \lambda \alpha k \mathbf{A} \vec{1})^\mathrm{T} \vec{\psi}\\
\text{s.t.}~ \ & 0 \le \psi_{i} \le 1, \ i = 1,2,...,N;\\
& \vec{\psi}^\mathrm{T} \vec{1} = \alpha N.
\end{split}
\label{eq:final}
\end{equation}

The relaxed optimization problem has a quadratic objective and linear constraints, which can be efficiently
solved with existing optimization algorithms, such as the interior-point method \cite{nesterov1994interior, wachter2006implementation}. 
Having acquired the solution $\vec{\psi}$ in $ [0,1] $, we regard each element of $\vec{\psi}$ as the confidence of the point to be selected. Points with top-$\alpha$ confidence are preserved while the rest are discarded for simplification.  

In order to speed up the algorithm, instead of constructing a graph over the entire point cloud and optimize the objective, we divide the point cloud into cubes and process each cube separately.
The size of a cube, i.e., the number of points in the cube, decides the efficiency of the algorithm.
While smaller size leads to faster implementation, larger size controls global information better.
In our experiments, we empirically constrain the size to the range [3000, 8000]. 

Further, artificial contours may occur along the boundary between two cubes due to the separate processing. In order to avoid this, for each cube $ \mathbb{C} $, we construct a graph over a larger cube that encompasses $ \mathbb{C} $, and compute the corresponding $ \mathbf{F} $ therein. The larger cube is able to contain all the $k$ nearest neighbors of the points in $ \mathbb{C} $, breaking the bound of $ \mathbb{C} $ and thus avoiding artificial contours. Then we resample $ \mathbf{F} $ to acquire that of the points in $ \mathbb{C} $.

Note that we simultaneously resample $\mathbf{A}$ when we resample $\mathbf{F}$. Then the degree of each point on the artificial block boundary will be less than $k$, i.e., the degree within the block of each point $\vec{x}_i$ on the block boundary $d_i<k$,  which contradicts our assumption. In order to address this issue, we complement $k-d_i$ for each point by adding it to the diagonal element of the binary adjacency matrix, i.e., $A_{i,i}$, thus avoiding blocking artifacts.

\section{Experimental Results}
\label{sec:results}
\begin{figure*}
	\centering
	\subfigure[Original]{
		\includegraphics[width=0.18\textwidth]{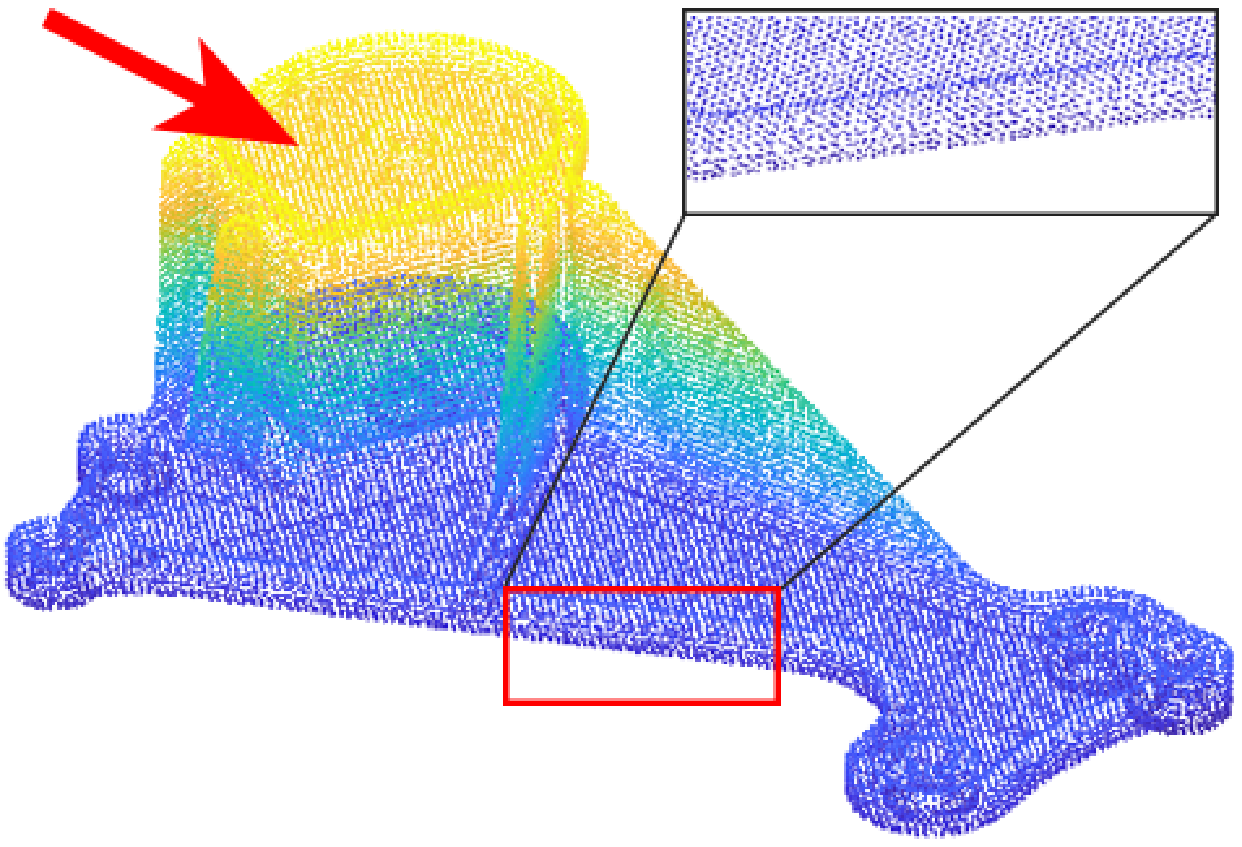}}
	\subfigure[Uniform]{
		\includegraphics[width=0.18\textwidth]{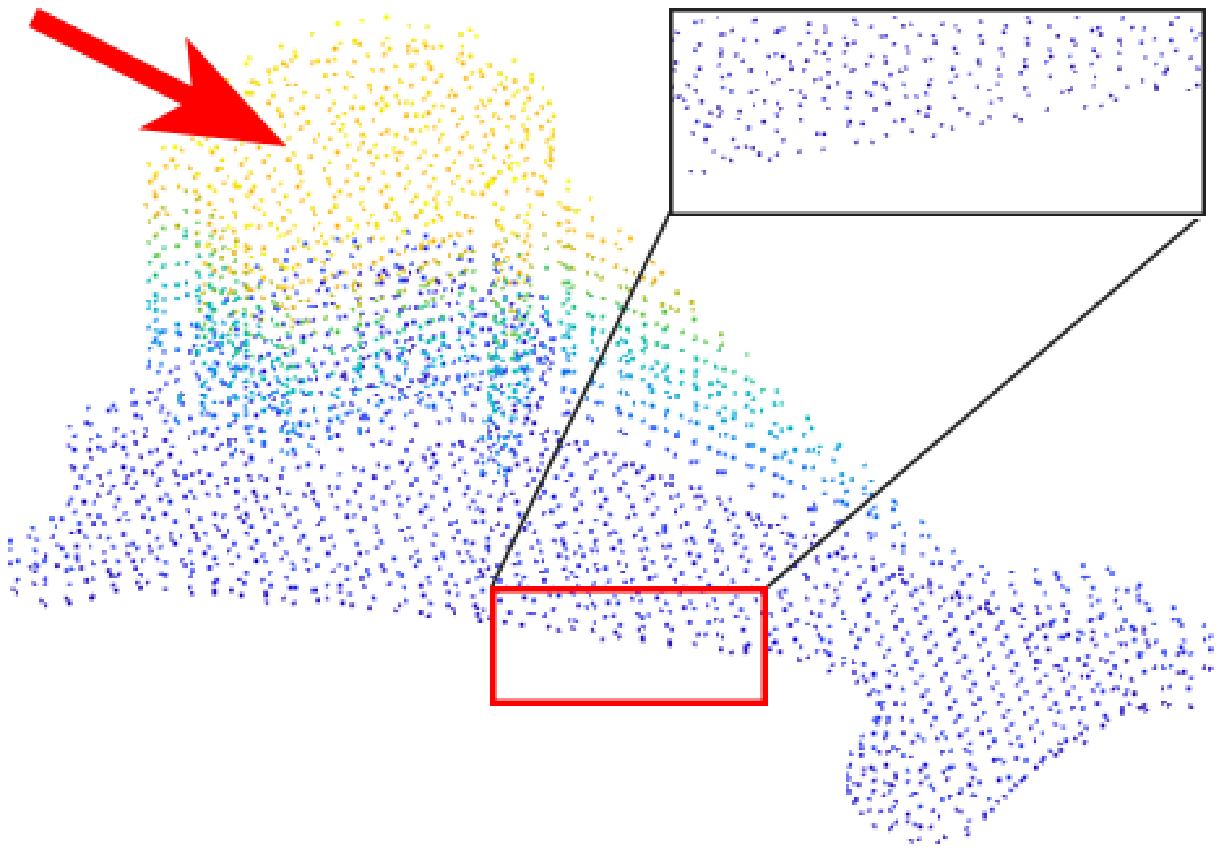}}
	\subfigure[Mean Curve]{
		\includegraphics[width=0.18\textwidth]{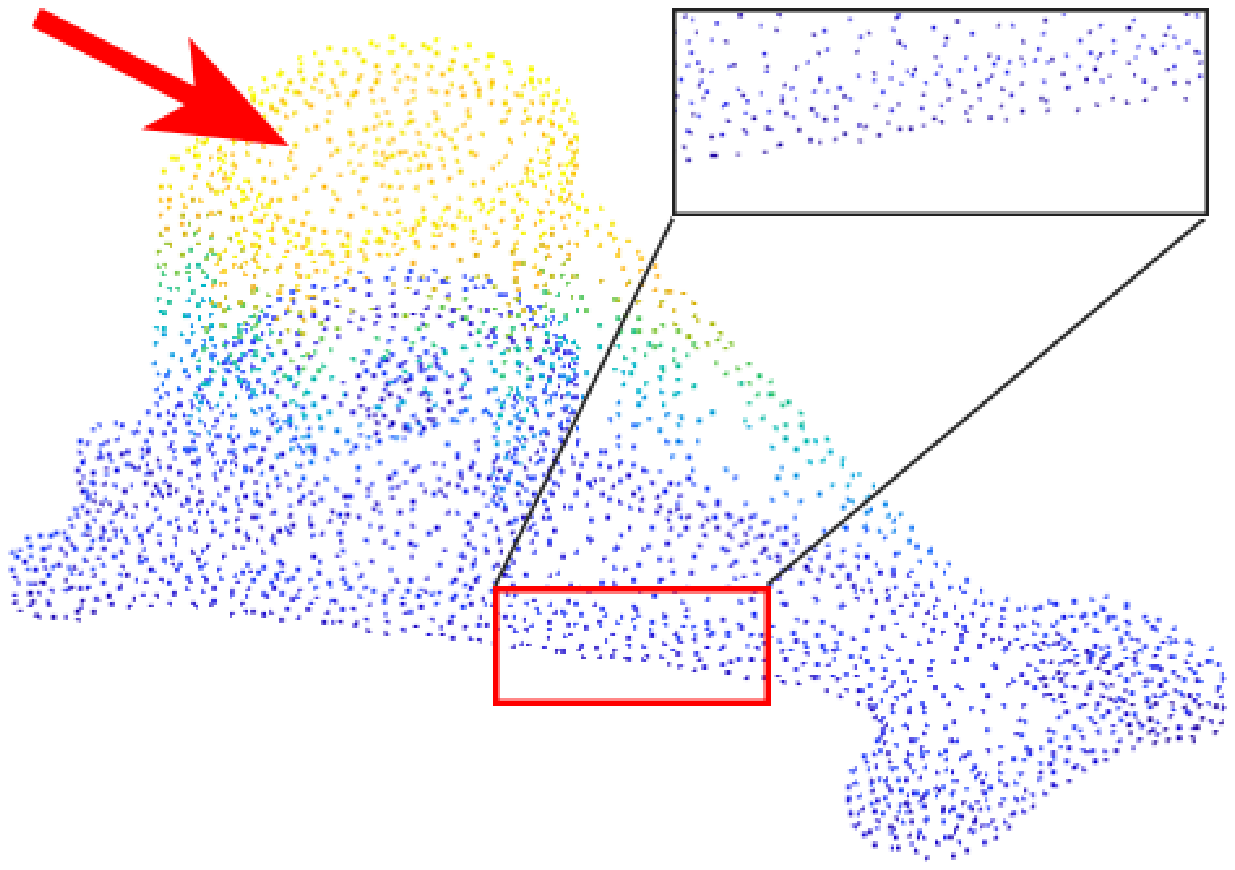}}	
	\subfigure[Contour]{
		\includegraphics[width=0.18\textwidth]{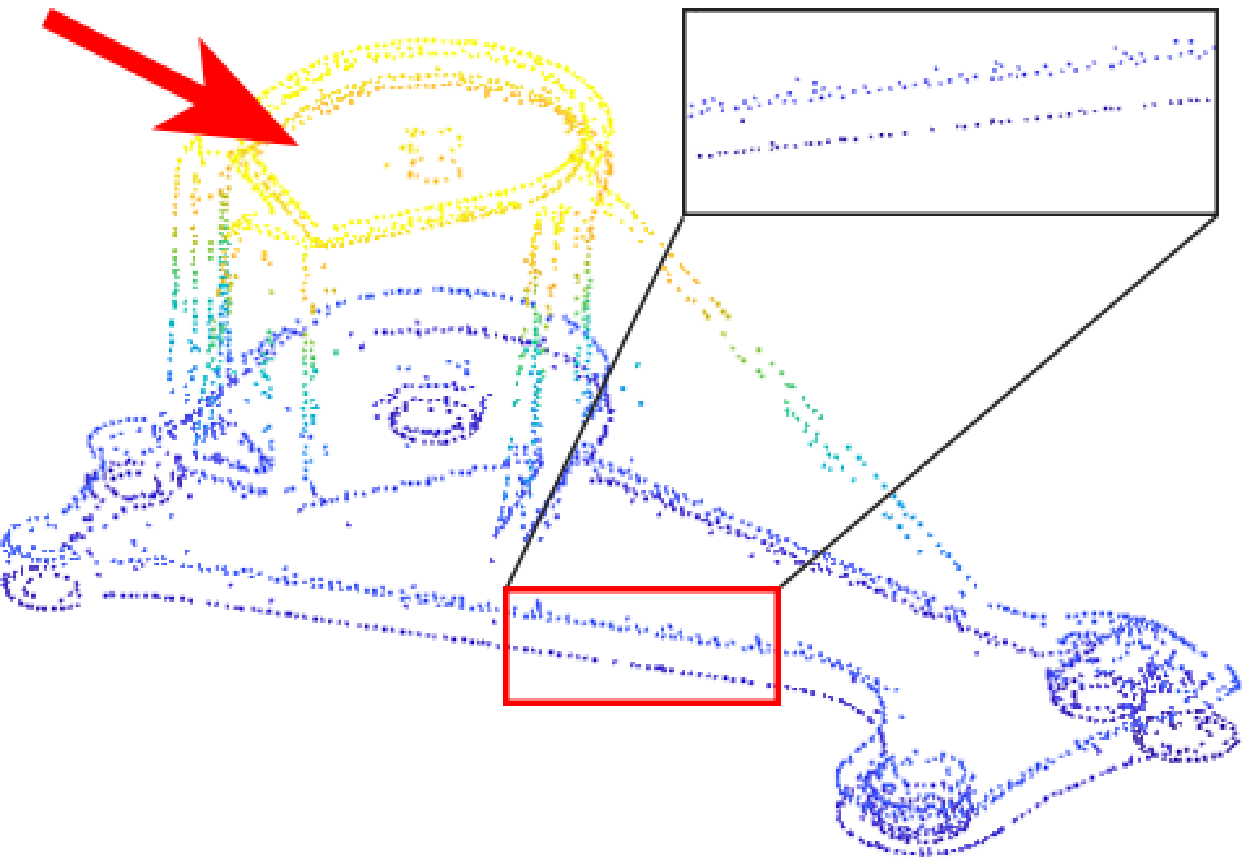}}
	\subfigure[Proposed]{
		\includegraphics[width=0.18\textwidth]{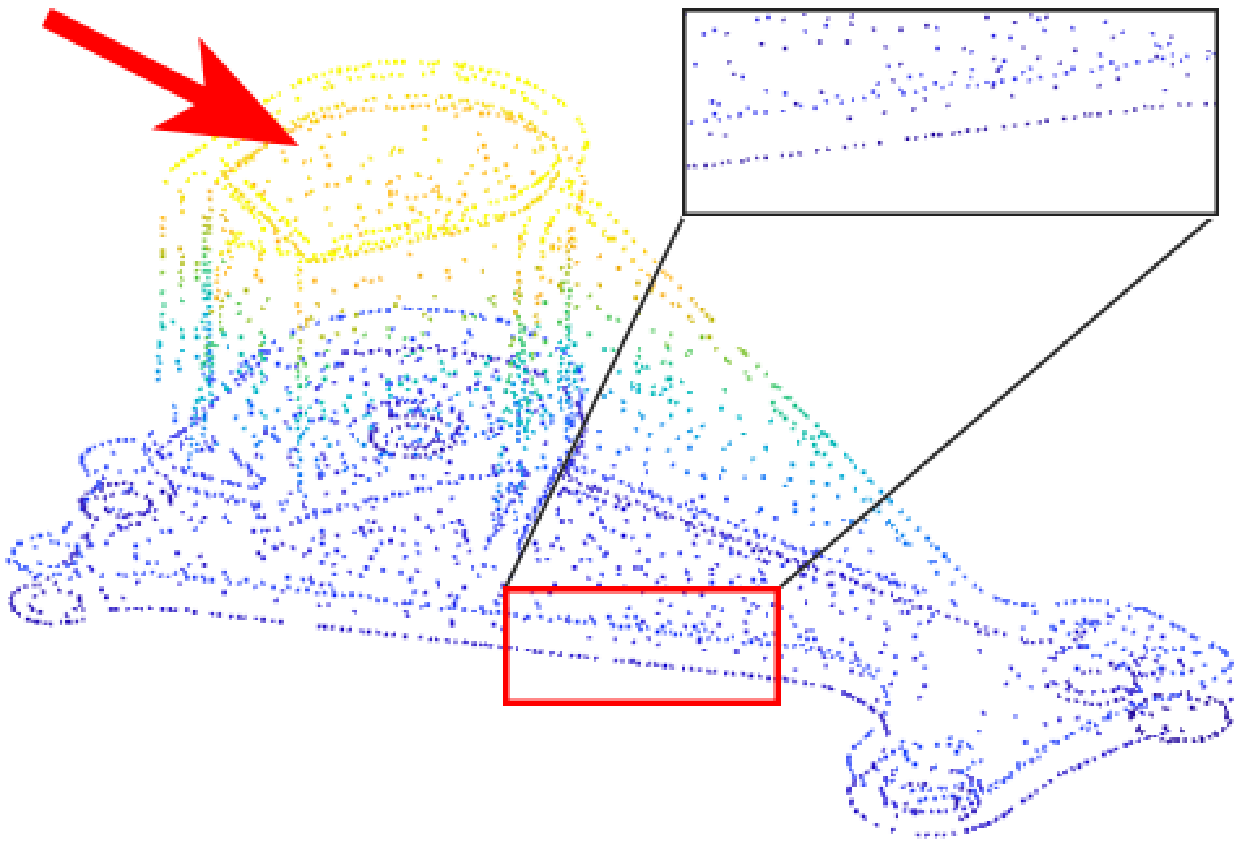}}
    \vspace{-0.1in}
	\caption{Simplification results for \textit{Daratech} with simplification rate 10\%. (b) loses much information such as edges. (c) preserves points around edges but is nonuniform (it looks uniform due to the fact that edges in \textit{Daratech} is close to each other. (d) preserves the contour effectively but keeps nearly no points on the smooth surfaces. (e) plays a good trade-off between the contour and uniformity. Please zoom in for more details.}
	\label{fig:daratech}
\end{figure*}

\begin{figure*}
	\centering
	\subfigure[Original]{
		\includegraphics[width=0.18\textwidth]{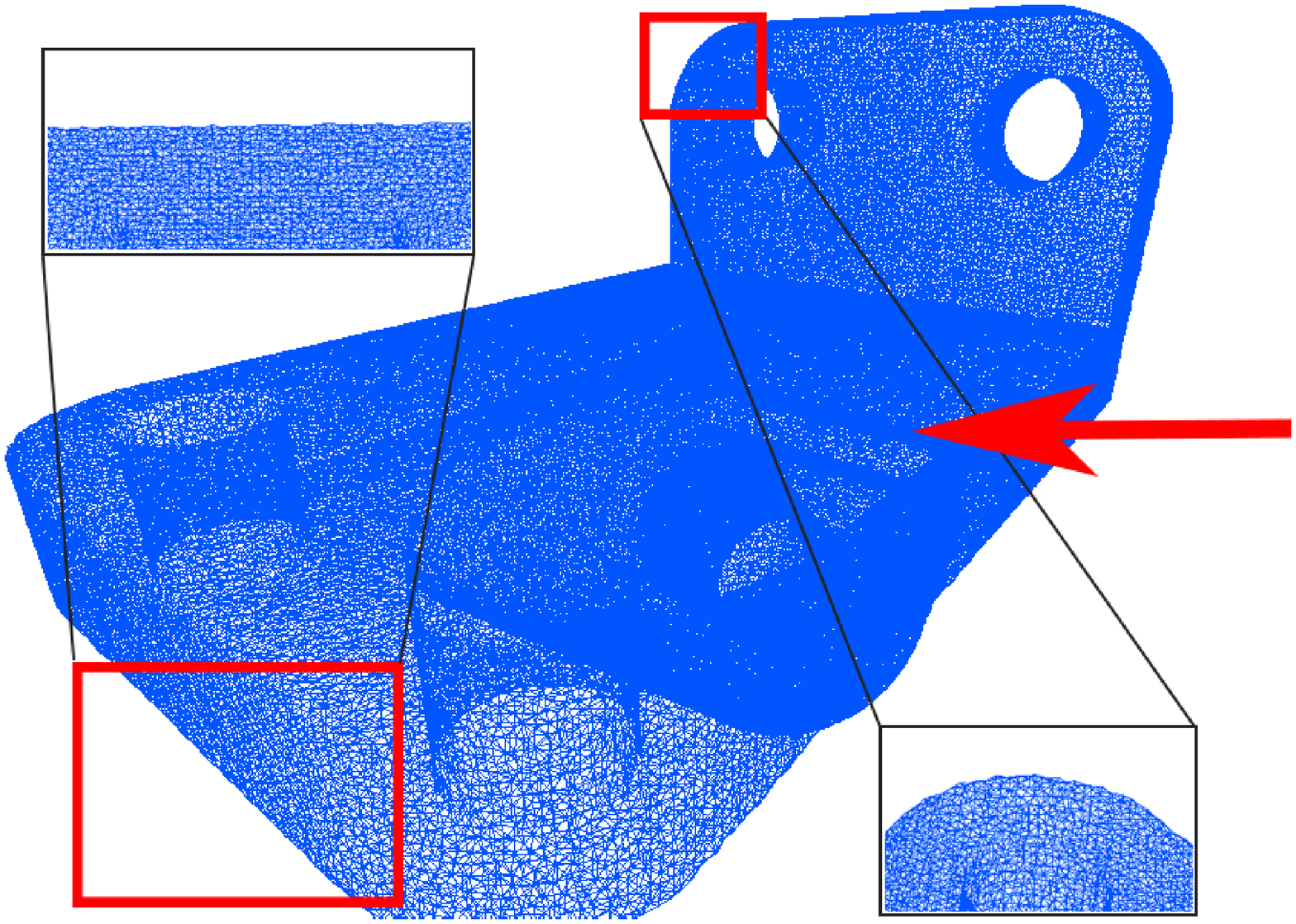}}
	\subfigure[Uniform]{
		\includegraphics[width=0.18\textwidth]{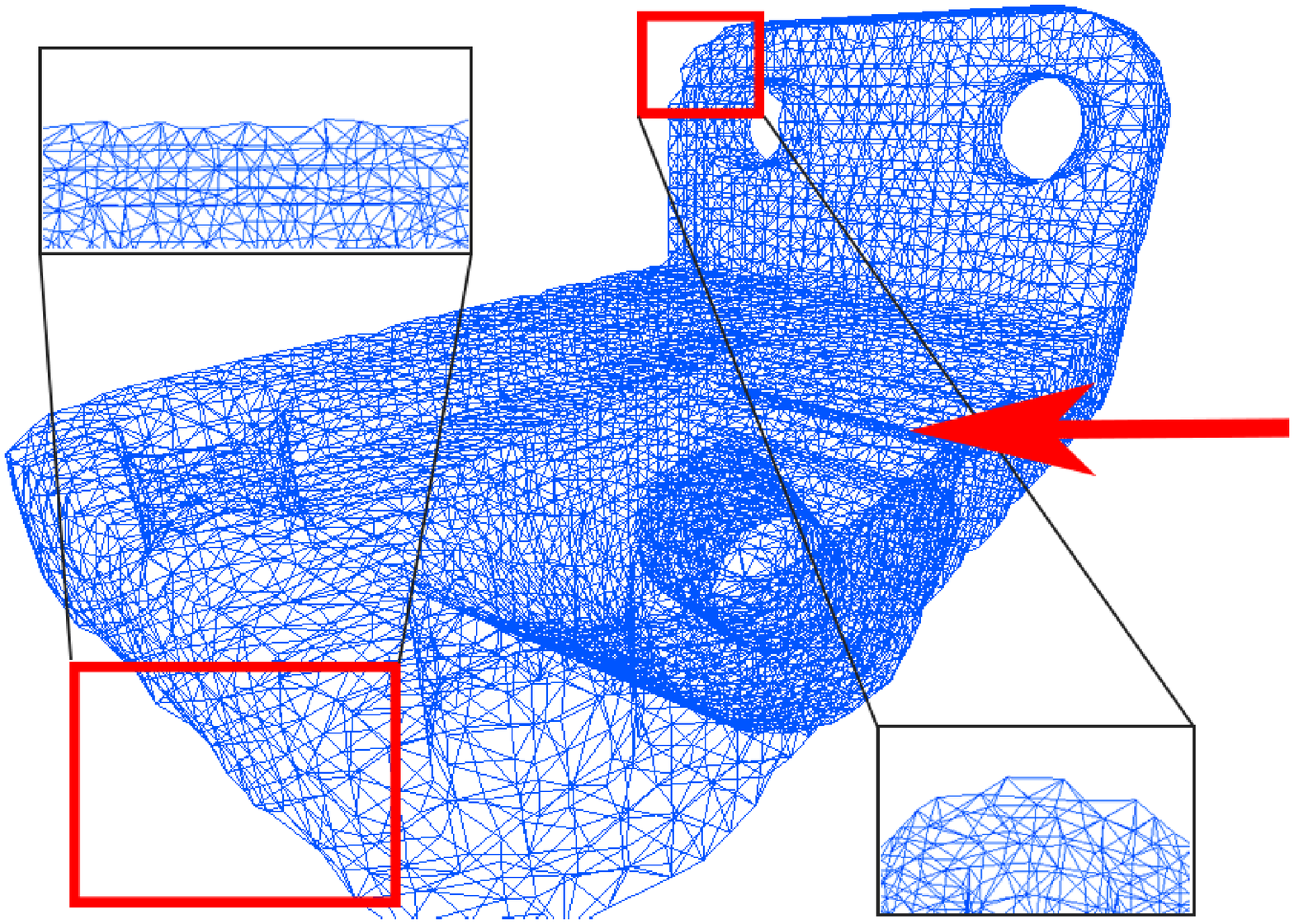}}
	\subfigure[Mean Curve]{
		\includegraphics[width=0.18\textwidth]{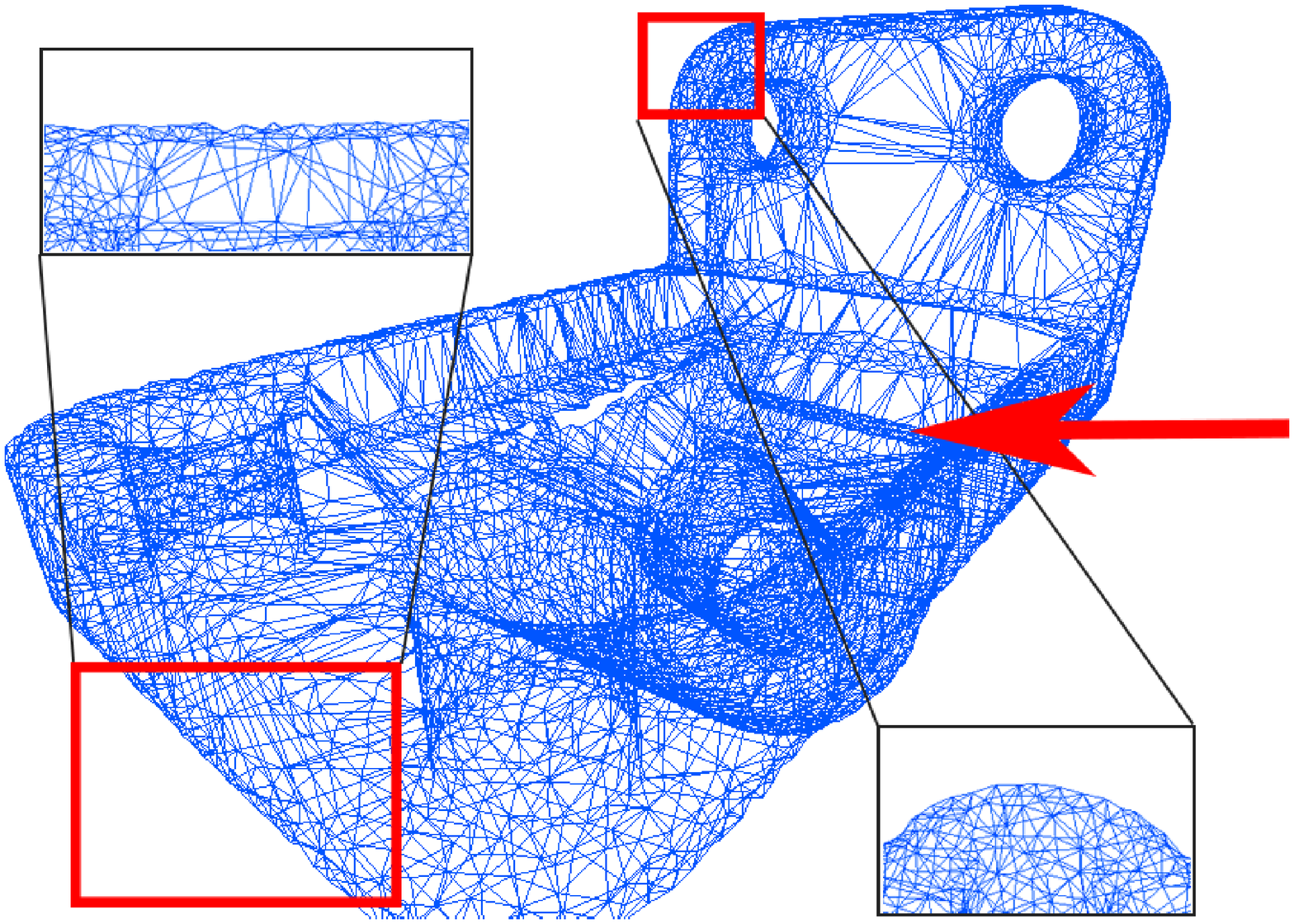}}
	\subfigure[Contour]{
		\includegraphics[width=0.18\textwidth]{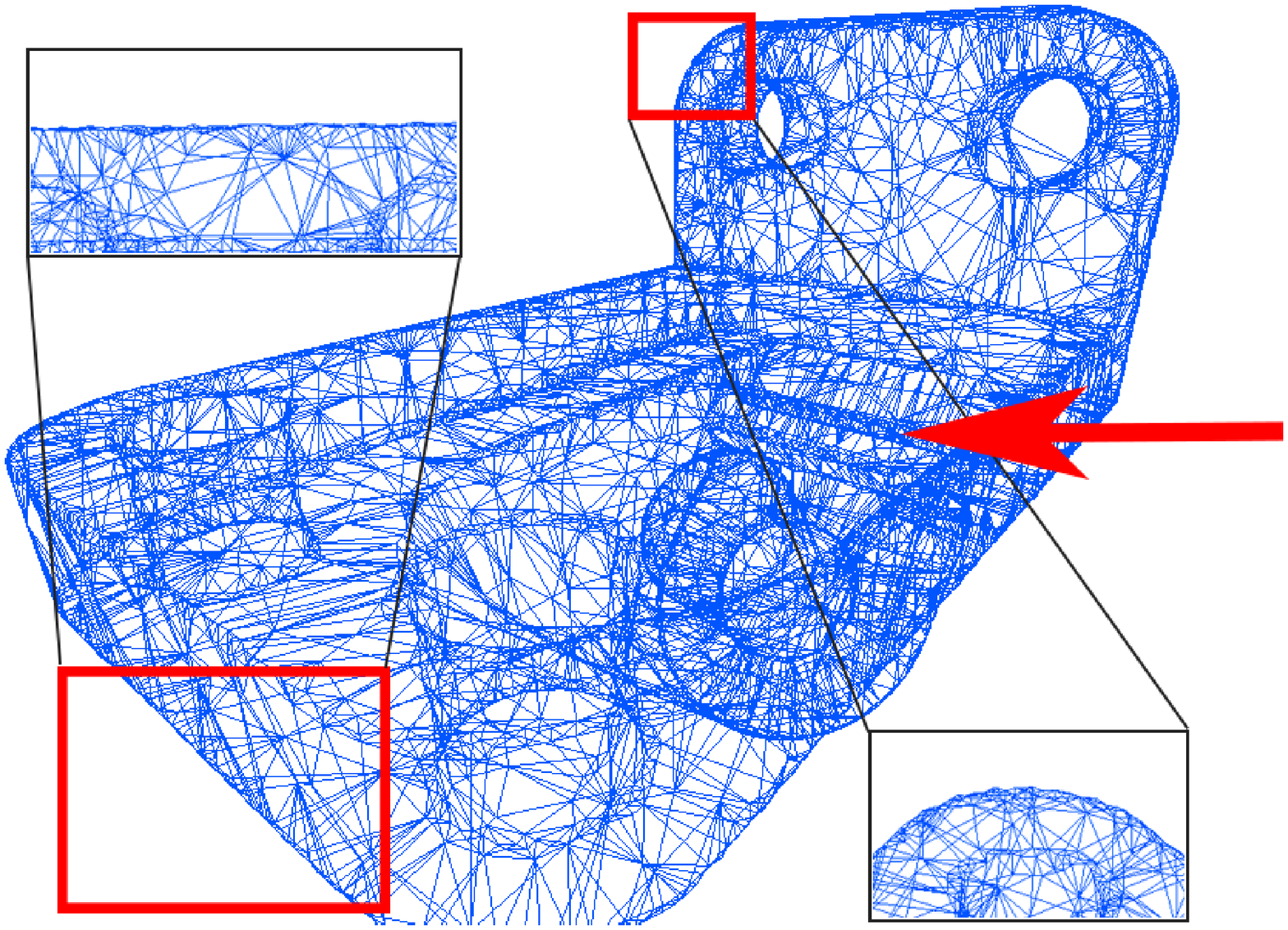}}
	\subfigure[Proposed]{
		\includegraphics[width=0.18\textwidth]{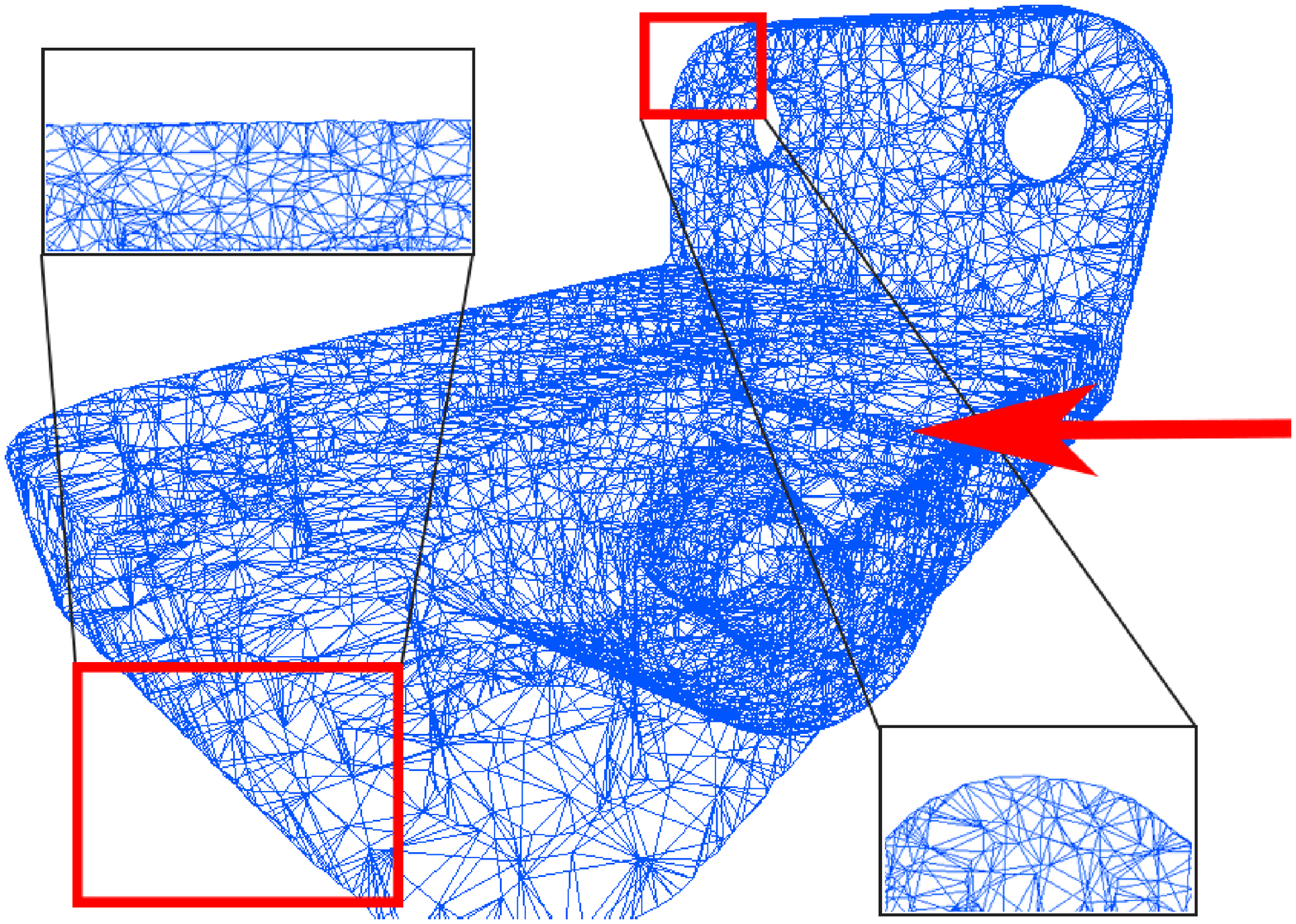}}
    \vspace{-0.1in}	
	\caption{Conversion of simplification results to meshes for \textit{Anchor} with simplification rate 10\%. (b) fails to restore edges accurately due to neglecting the contour information. (c) preserves the points around edges but neglects points on the surface, leading to a big hole as pointed out by the red arrow. (d) leads to triangles of extremely uneven size in the converted mesh. (e) exhibits a good trade-off. Please zoom in for more details.}
	\label{fig:anchor}
\end{figure*}

\begin{figure*}
	\centering
	\subfigure[$\lambda=10^{-1}$]{
		\includegraphics[width=0.18\textwidth]{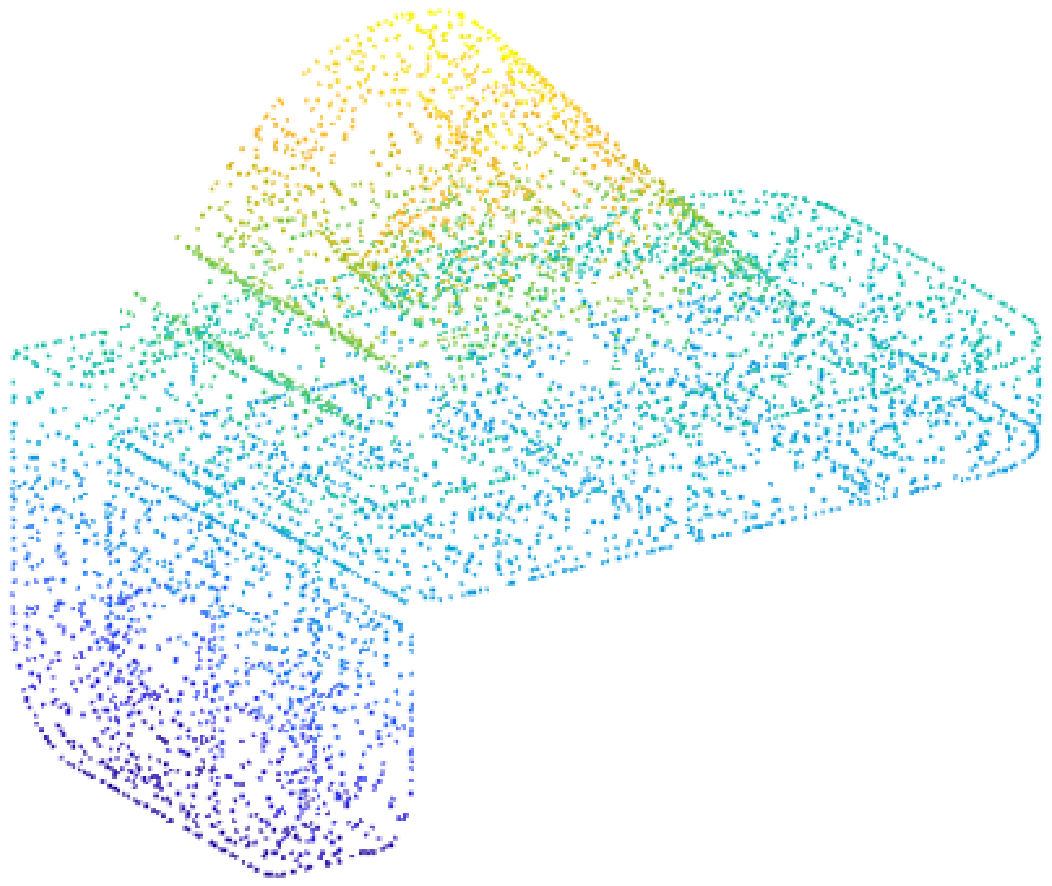}}
	\subfigure[$\lambda=10^{-3}$]{
		\includegraphics[width=0.18\textwidth]{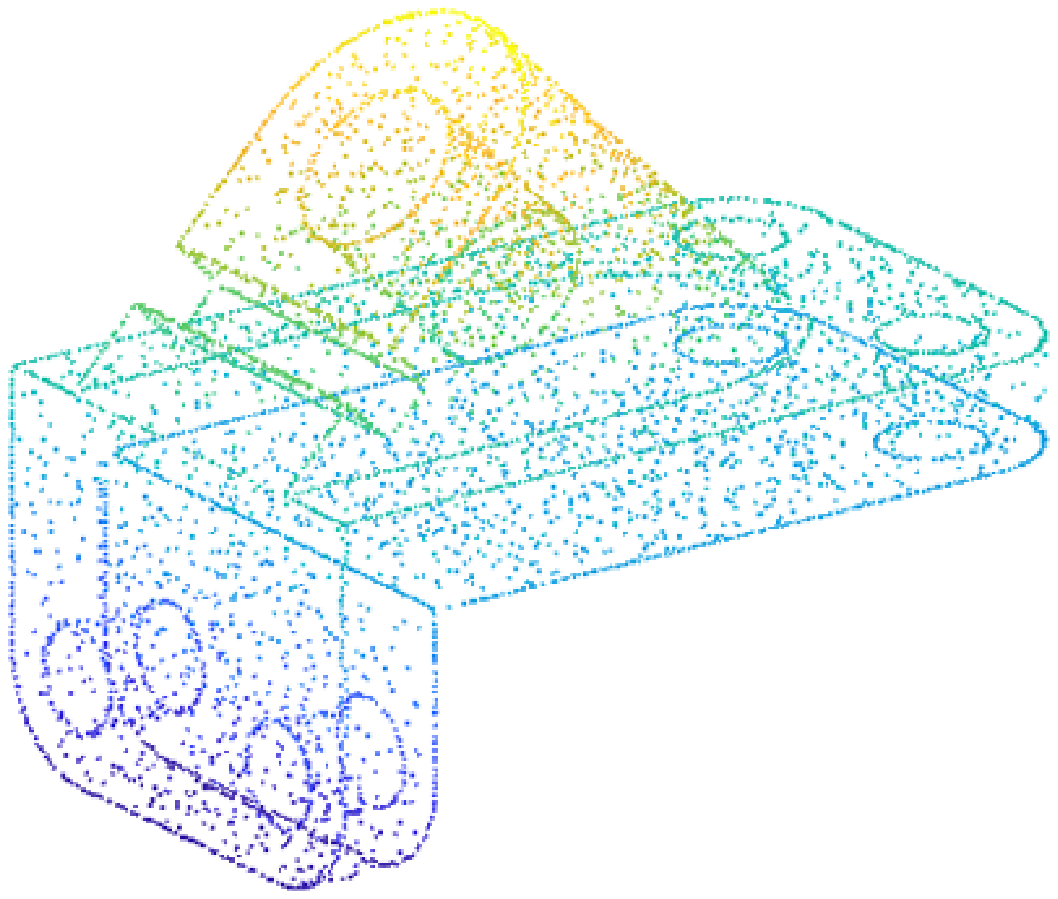}}
	\subfigure[$\lambda=10^{-5}$]{
		\includegraphics[width=0.18\textwidth]{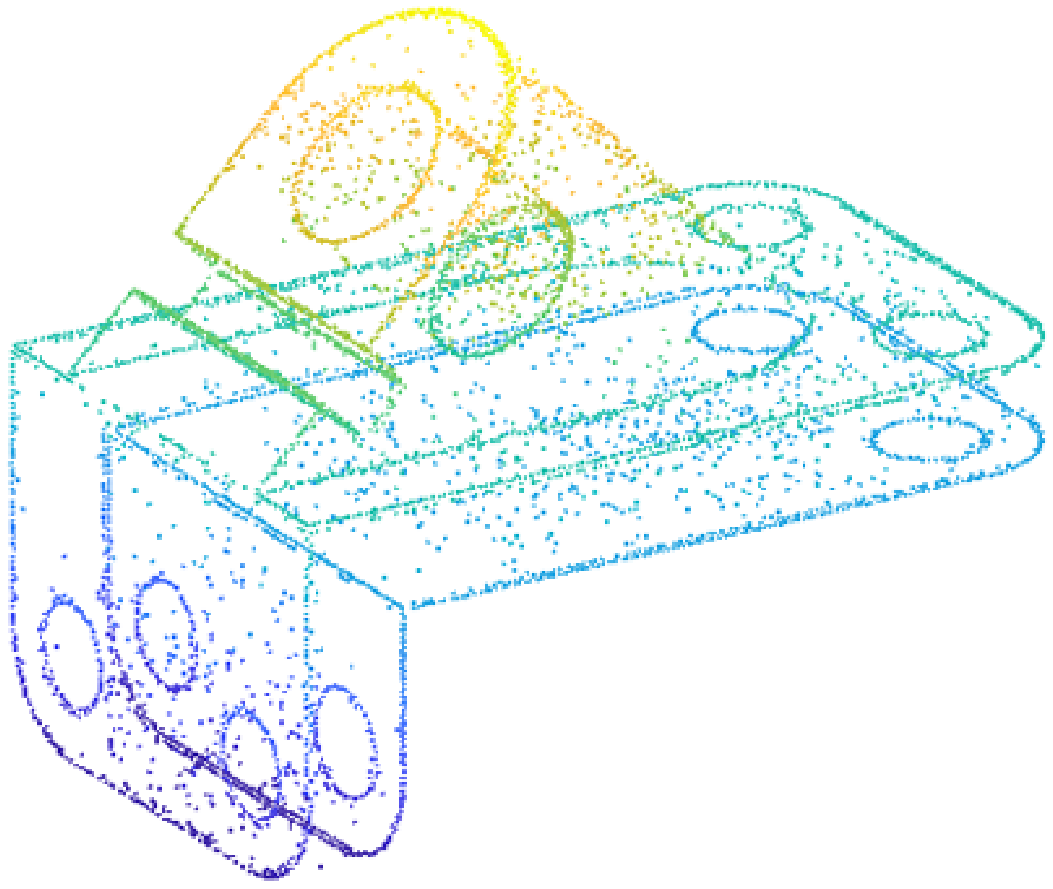}}
    \vspace{-0.1in}	
	\caption{Simplification results for \textit{Anchor} with different settings of $\lambda$ using the proposed method. Please zoom in for more details.}
	\label{fig:lambda}
\end{figure*}

\begin{figure*}
	\centering
	\subfigure[Transformed]{
		\includegraphics[width=0.22\textwidth]{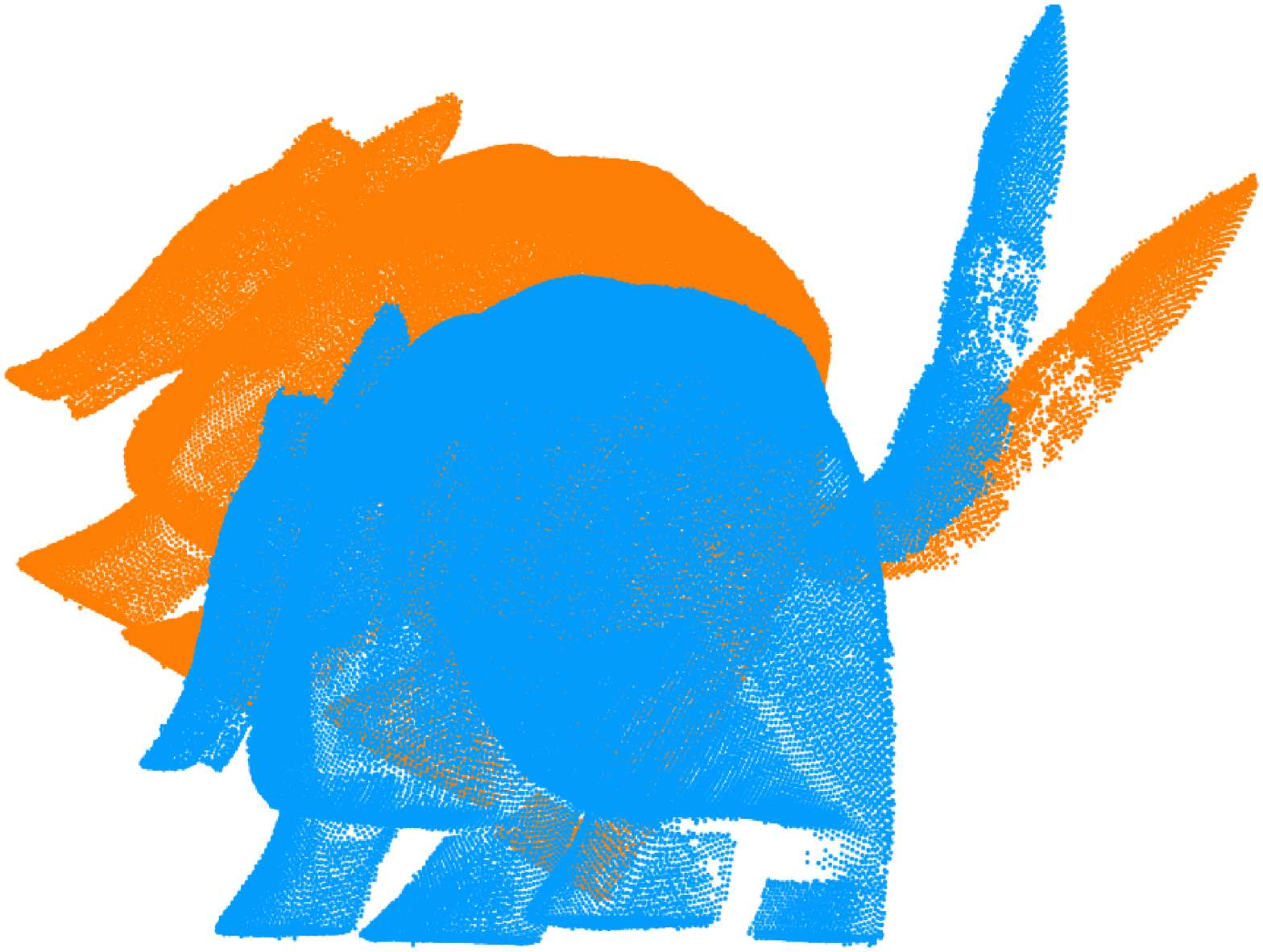}}
	\subfigure[Original]{
		\includegraphics[width=0.18\textwidth]{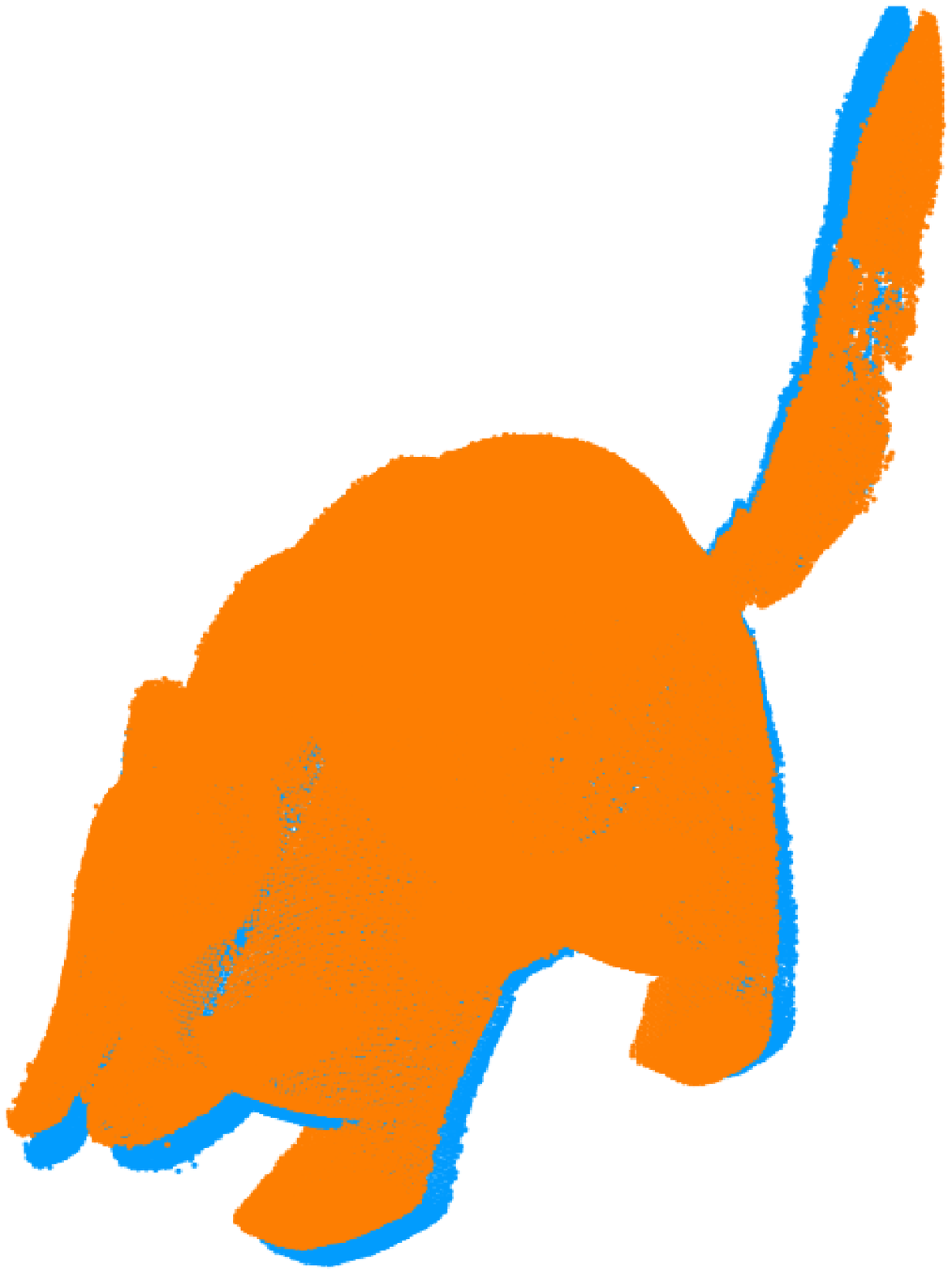}}
	\subfigure[Uniform]{
		\includegraphics[width=0.18\textwidth]{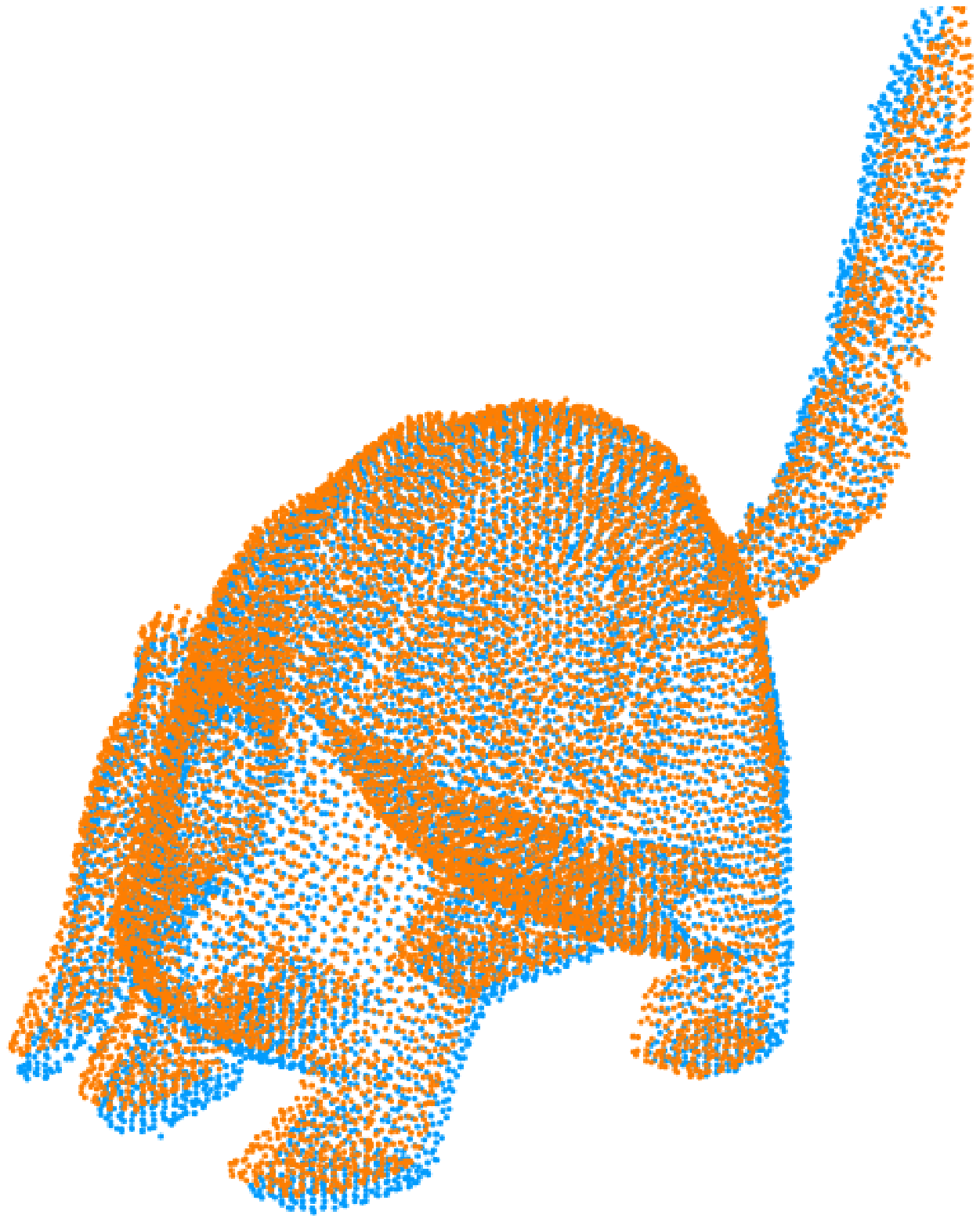}}
	\subfigure[Proposed]{
		\includegraphics[width=0.18\textwidth]{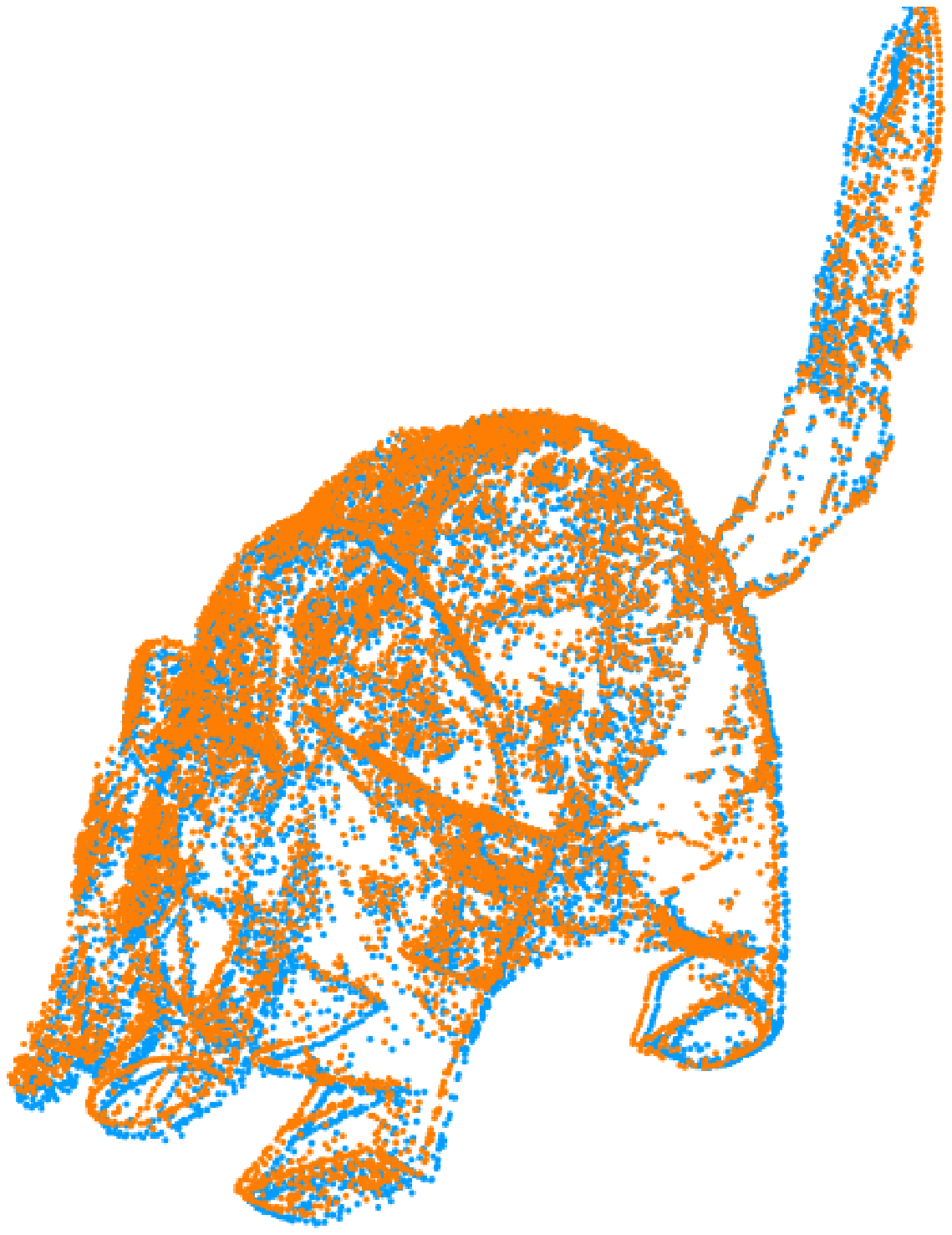}}
    \vspace{-0.1in}
	\caption{Registration results for \textit{Armadillo}. (a) shows the point clouds before registration. (b)(c)(d) are registration results from different simplification methods. (d) enhances the contour information while maintaining the uniform density, thus leading to the best registration result. Please zoom in for more details.}
	\label{fig:armadillo}
\end{figure*}

\subsection{Experimental Setup} 
In order to evaluate the proposed method, we compare with three competitive approaches: the mean-curve based feature-preserving simplification in \cite{yang2015feature}, the graph-based contour-extracted resampling in \cite{chen2018fast}, and the uniform sampling method using the voxel-grid in PCL \cite{rusu2011pcl}.
We test on several point clouds, including
\textit{Daratech}, \textit{Anchor}, \textit{Armadillo}, \textit{Shutter} \cite{mattei2017point}, and \textit{Hand} \footnote{\url{https://www.cc.gatech.edu/projects/large_models/}}.

\subsection{Experimental Results} 
\textbf{Visualization} We deploy the proposed simplification method to efficiently visualize large-scale point clouds. Two representative results are presented in Fig. \ref{fig:daratech} and Fig. \ref{fig:anchor}. 
The former is demonstrated in the format of point clouds, while the latter is converted to meshes for applications such as computer graphics where meshes are preferable due to the available topology. We observe that uniform-sampling \cite{rusu2011pcl} leads to results with defects along corners, as demonstrated in red boxes in Fig. \ref{fig:daratech} \& \ref{fig:anchor}, because the method neglects sharp features of the point cloud. 
The feature-aware methods \cite{chen2018fast, yang2015feature} preserve sharp features well but fail to keep the uniformity of the point cloud, as pointed with red arrows in Fig. \ref{fig:daratech} \& \ref{fig:anchor}. In contrast, the proposed method not only preserves points with sharp feature (e.g. points on the contours), but also keeps the uniformity to some extent for better visualization and mesh conversion.

Next we evaluate the power of the parameter $\lambda$ in controlling the uniformity of the point cloud in Fig. \ref{fig:lambda}.
As mentioned, $\lambda$ is the weight of the loss in uniformity in (\ref{eq:final}).
A smaller $\lambda$ leads to less constraint on the uniformity and thus preserves more sharp features. The adjustable parameter $\lambda$ enables users to conveniently control the uniformity of the simplified point cloud.

\textbf{Application to Registration} We apply our simplification method to accurate registration of large point clouds.
We intentionally shift and rotate the original point cloud to obtain the transformed point cloud.
The universal ICP algorithm \cite{besl1992method} is employed to register the simplified original and transformed point clouds.
As presented in the visualization experiments, feature-aware methods \cite{chen2018fast, yang2015feature} preserve sharp features well but lead to extremely nonuniform point clouds, which is often unsuitable for the subsequent processing such as the aforementioned mesh conversion. Thus these two methods are not applied in registration. We compare with the original-sized and uniformly resampled point clouds, which are two most common strategies used in point cloud registration.

We adopt the root mean square error $\mathbf{RMSE}$ as the evaluation metric for registration.
Specially, $ \mathbf{RMSE} = \sqrt{\frac 1 N \sum_{i=1}^N \|\hat{\vec{x}}_i - \vec{x}_i \|_2^2 } $,
where $\hat{\vec{x}}_i \in \mathbb{R}^3 $ is the coordinate of the $i$-th point for the simplified point cloud, while $\vec{x}_i$ is that of the ground truth.

The quantitative results are listed in Tab.~ \ref{tab:comparison} and one of the visual results is shown in Fig. \ref{fig:armadillo}.
We set the simplification rate as 10\% for \textit{Anchor} and \textit{Armadillo} (55,799 and 99,416 points respectively), and 5\% for large \textit{Shutter} and \textit{Hand} (291,220 and 327,323 points respectively).
We see that the proposed method outperforms the other methods, which is consistent with the visual results in Fig. \ref{fig:armadillo} and validates the effectiveness of our method.  

\begin{table}[!htbp]
\centering
\caption{Quantitative Results of Registration in \textbf{RMSE}}
\vspace{0.1in}
\begin{tabular}{|c|c|c|c|c|}
    \hline
    \textbf{Method} & \textit{Anchor} & \textit{Armadillo} & \textit{Shutter} & \textit{Hand}\\
    \hline
    {Original} & {0.8325} & {0.0155} & {0.0248} & {0.1575}\\
    \hline
    {Uniform} & {0.9348} & {0.0155} & {0.0248} & {0.1588}\\
    \hline
    {Proposed} & \textbf{0.5008} & \textbf{0.0124} & \textbf{0.0205} & \textbf{0.1508}\\
    \hline
\end{tabular}
\label{tab:comparison}
\end{table}

\section{Conclusion}
\label{sec:conclude}
Leveraging on graph signal processing, we propose an efficient point cloud simplification method, which strikes a balance between preserving sharp features and keeping uniform density. A concise formulation is presented based on graph filters, which offers an adjustable parameter to control the uniform density conveniently according to the applications. Experimental results demonstrate the effectiveness of the proposed method and its application to point cloud registration. 

\begin{footnotesize}
\bibliographystyle{IEEEbib}

\end{footnotesize}

\end{document}